\def\year{2020}\relax
\documentclass[letterpaper]{article} % DO NOT CHANGE THIS
\usepackage{aaai20}  % DO NOT CHANGE THIS
\usepackage{times}  % DO NOT CHANGE THIS
\usepackage{helvet} % DO NOT CHANGE THIS
\usepackage{courier}  % DO NOT CHANGE THIS
\usepackage[hyphens]{url}  % DO NOT CHANGE THIS
\usepackage{graphicx} % DO NOT CHANGE THIS
\usepackage{graphicx}  % DO NOT CHANGE THIS
\usepackage{times}
\usepackage{latexsym}
\usepackage{url}
\usepackage{multirow}
\usepackage{array}
\usepackage{amsmath}
\usepackage{graphicx} 
\usepackage{colortbl}
\definecolor{mygray}{gray}{.80}
\usepackage{subfigure}
\newcommand{\annotator}[0]{\includegraphics[width=0.03\textwidth]{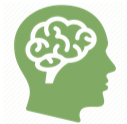}} 
\newcommand{\checker}[0]{\includegraphics[width=0.03\textwidth]{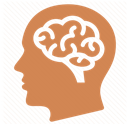}} 
\newcommand{\machine}[0]{\includegraphics[width=0.03\textwidth]{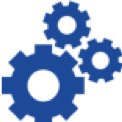}} 
\newcommand{\citet}[1]{\citeauthor{#1}~\shortcite{#1}}
\newcommand{\citep}{\cite}

\newcolumntype{T}{>{\tiny}l} % define a new column type for \tiny
\title{ReCO: A Large Scale Chinese\\ Reading Comprehension Dataset on Opinion}
\author{Bingning Wang, Ting Yao, Qi Zhang\\ \Large \textbf{ Jingfang Xu, Xiaochuan Wang}\\ 
	Sogou Inc.\\ %If you have multiple authors and 
	Beijing, 100084, China\\
	\{wangbingning,yaoting,qizhang,xujingfang,wxc\}@sogou-inc.com 
}
\begin{document}
	
	\maketitle
	
	\begin{abstract}
		This paper presents the ReCO, a human-curated Chinese \underline{Re}ading \underline{C}omprehension dataset on \underline{O}pinion. The questions in ReCO are opinion based queries issued to commercial search engine. The passages are provided by the crowdworkers who extract the support snippet from the retrieved documents. Finally, an abstractive yes/no/uncertain answer was given by the crowdworkers. The release of ReCO consists of 300k questions that to our knowledge is the largest in Chinese reading comprehension. A prominent characteristic of ReCO is that in addition to the original context paragraph, we also provided the support evidence that could be directly used to answer the question. Quality analysis demonstrates the challenge of ReCO that it requires various types of reasoning skills such as causal inference, logical reasoning, etc. Current QA models that perform very well on many question answering problems, such as BERT \citep{devlin2018bert}, only achieves 77\% accuracy on this dataset, a large margin behind humans nearly 92\% performance, indicating ReCO present a good challenge for machine reading comprehension. The codes, dataset and leaderboard will be freely available at \url{https://github.com/benywon/ReCO}.

	\end{abstract}
	
	\section{Introduction}\label{intro}
	
	Machine reading comprehension (MRC), the ability to read the text and answer questions, has become one of the mainstreams in current natural language understanding (NLU) researches. Compared to other types of QA, MRC provided with only one document so the statistical information such as the number of answer occurrences could not be utilized, thus it requires a deeper understanding of the text. MRC has become an important part in many natural language processing applications, such as information retrieval \citep{Nishida2018RetrieveandReadML}, event extraction \citep{Ramamoorthy2018AnAS} and relation extraction \citep{Levy2017ZeroShotRE}.

	One of the major contributions of the dramatic progress in MRC is the development of large scale corpus. Since the release of primal MCTest \citep{richardson2013mctest}, a great amount of datasets have been proposed, such as SQuAD \citep{rajpurkar2016squad,Rajpurkar2018KnowWY}, CNN/Daily Mail \citep{Hermann2015TeachingMT}, RACE \citep{Lai2017RACELR}, NarrativeQA \citep{Kocisk2018TheNR}, etc. Based on these large scale datasets, a lot of deep learning based models have been built, such as BiDAF \citep{Seo2016BidirectionalAF}, QANet \citep{Yu2018QANetCL}, etc. These models behave very well in MRC and some of them even surpass human performance.
	
	However, despite the various types and relatively large scale, we found there are two main challenges previous MRC datasets has not been addressed:

	\textbf{ 1) The MRC context in most previous datasets are limit to the relatively long document or paragraph, which contains much irrelevant information to the question.} Therefore, the \textit{comprehension} process is sometimes reduced to the \textit{retrieval} process \citep{sugawara2018makes}, an MRC system could perform very well by merely finding the relevant sentences in a paragraph. For example, in SQuAD and NewsQA, the model's performance did not downgrade when only provided the sentence containing the ground truth answer \citep{Weissenborn2017MakingNQ,min2018efficient}. In NarrativeQA \citep{Kocisk2018TheNR} or NaturalQuestions \citep{kwiatkowski2019natural} where the context passage is very long, the answer selection became the dominant factor for final result \citep{alberti2019bert,kwiatkowski2019natural}. The answer generation of MRC, which requires deep understanding of the text, has not been throughout evaluated.
	
	\textbf{2) Most previous MRC datasets are focus on factoid questions} such as \textit{who}, \textit{when}, \textit{where}, etc., so the candidate answers are limited to certain types such as \textit{person}, \textit{time}, \textit{position}. Therefore, this kind of question does not require a complex understanding of language but merely recognizing the entity type could solve it properly. \citet{sugawara2018makes} show that only using the first several tokens of the questions could achieve a significant improvement over random selection in many MRC datasets. This makes the reasoning process of machine learning methods built upon these datasets questionable \citep{Jia2017AdversarialEF}. 
	
	%     3) In most of the previous datasets, the domains of the document are limited to certain types, such as Wikipedia \citep{rajpurkar2016squad,kwiatkowski2019natural}, news articles \citep{Hermann2015TeachingMT} or children stories \citep{Hill2016TheGP,richardson2013mctest}, which elude understanding the stylistic feature of text and hard to test the generalization ability of existing models.
	
\begin{figure}
		\centering
		\includegraphics[width=0.95\linewidth]{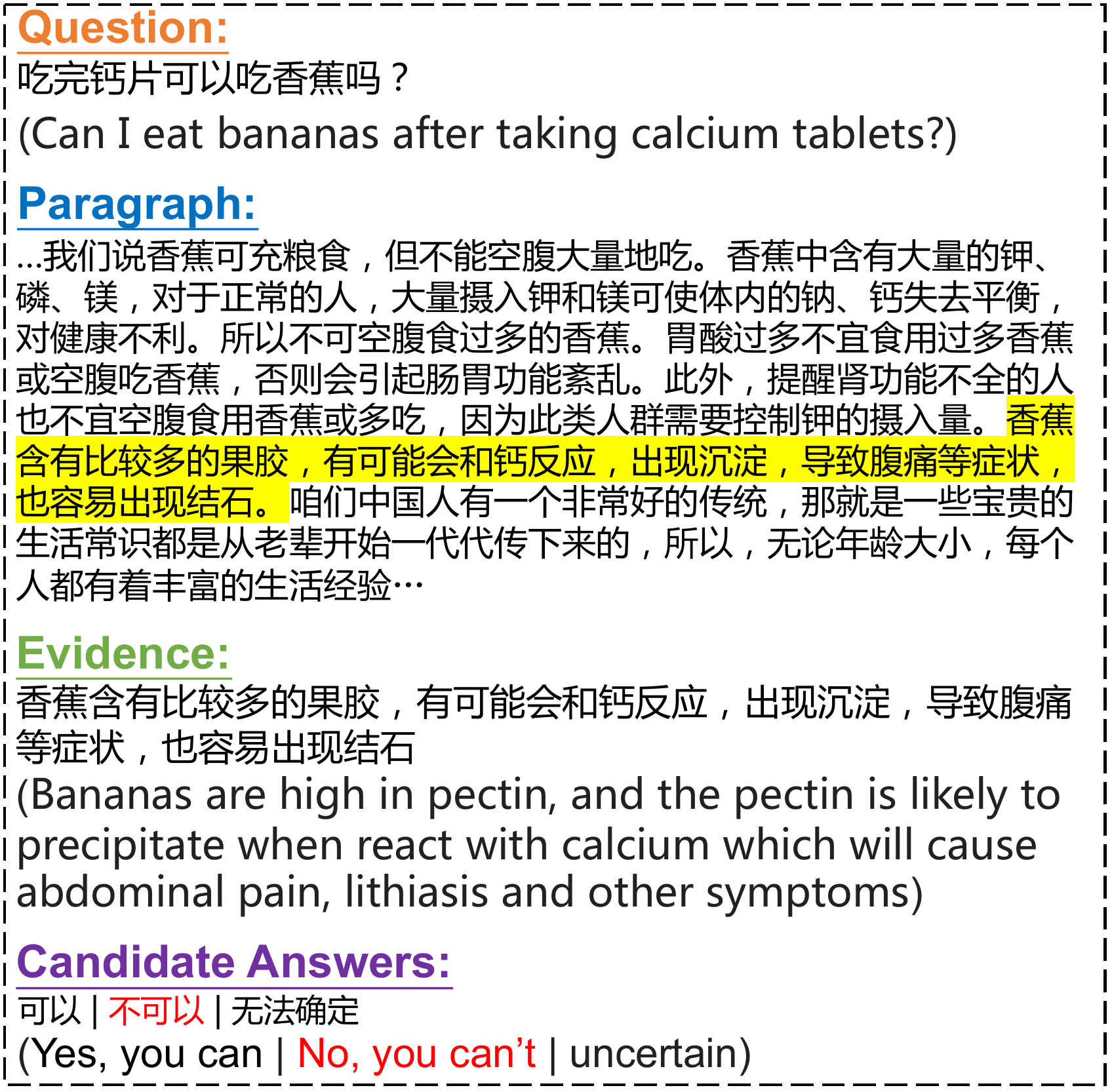}
		\caption{An example of ReCO. The evidence is extracted from the paragraph, which contains less irrelvant information to answer the question. The evidence may or may not consist of consecutive sentences in the paragraph.}\label{fig_example}
	\end{figure}

	In this paper, we present ReCO, a large scale human-curated Chinese reading comprehension dataset focusing on opinion questions. In ReCO, the questions are real-world queries issued to search engine\footnote{\url{https://www.sogou.com/}}. These queries are sampled and further filtered such that each query is a valid question and can be answered by yes/no/uncertain. Given the question, 10 retrieved documents are provided to the annotators, they were asked to select one document and extract the support evidence for that question. 
	Finally, three candidate answers were given by the annotator: a positive one like \textit{Yes}, a negative one like \textit{No}, and an undefined one if the question could not be answered with the given documents. An example is shown in Figure \ref{fig_example}.
	
	Compared with previous MRC datasets, there are three main characteristics of the ReCO: First of all, in addition to the original document, we also provided the support evidence for the question. We do this because 1) Some documents may directly contain the answers with yes/no, where the answer could be trivially answered without understanding the subsequent evidence. 2) removing the irrelevant information could bypassing the answer selection error and concentrates the ReCO on the inference process of MRC. Data analysis shows that a large amount of ReCO questions require deep reasoning skills of text such as causal inference, logical reasoning, etc. (3), the $ \langle \text{\textit{paragraph, evidence}}\rangle $ could be utilized for further NLP applications such as summarization or answer selection.
	
	Secondly, the questions in ReCO are opinion based real-world queries, which may be either factoid or non-factoid, and spanning many types of domains such as medical, health, etc. Besides, we use the search engine to obtain the passages which come from various resources, such as news articles, community QA or cyclopedia, etc. The diversity of questions and documents endows ReCO contains many aspects of world knowledge.
	
	Finally, ReCO is very large and high-quality: it contains 300k questions, to our best knowledge it is the largest human-annotated opinion based QA dataset. In addition to the large scale, we introduce a rigorous inspection strategy to ensure the quality of the data, this makes ReCO relatively hard requiring deep understanding of text.
	%     that it could only be answered by deep understanding of the text. 
	
	We applied several models on ReCO, including a modified version of BERT \citep{devlin2018bert} to fit the multiple-choice problem. The experimental results show that although it is a simple one-out-of-three problem, the best model only achieves 77\% accuracy compared to humans 92\%. The large gap between machine systems and humans indicates ReCO providing a good testbed for NLU systems.
	
	\begin{table*}
		\centering
		%        \small
		\begin{tabular}{lcccccc}
			\textbf{dataset} & \textbf{type} & \textbf{question source } & \textbf{passage source }                                                              &\textbf{ answer source }   & \textbf{datasize} & \textbf{question type} \\ \noalign{\hrule height 1.2pt}     
			SQuAD         & OQA       & human generated  & wiki passage                                                                 & human extracted  & 100k     & F       \\ \hline
			SearchQA       & OQA       & J! Archive       & search result                                                                & J! Archive       & 100k     & F       \\ \hline
			MCTest         & MC        & human generated  & story                                                                        & human generated  & 2,000    & NF       \\ \hline
			CNN/Daily Mail & CLOZE     & abstract summary & news                                                                         & entities         & 1.4m     & F       \\  \hline
			CBT & CLOZE     & children's book & children's book                    & entities         & 680k     & NF       \\  \hline
			MARCO          & OQA       & query logs       & search result                                                                & human generated  & 100k     & F/NF       \\ \hline
			NarrativeQA    & OQA       & human generated  & \begin{tabular}[c]{@{}c@{}}books and\\  movie scripts\end{tabular}           & human generated  & 44k      & NF       \\ \hline
			NaturalQuestions      & OQA       & queries logs     & wiki document                                                                & human extracted  & 350k     & F       \\ \hline
			DRCD      & OQA       & human generated      & wiki document                                                                & human extracted  & 30k     & F       \\ \hline
			DuReader       & OQA       & queries logs     & search result                                                                & human extracted  & 200k     & F/NF       \\ \hline \rowcolor{mygray}
			ReCO           & MC        & queries logs     & \begin{tabular}[c]{@{}c@{}}extraction from \\  search result\end{tabular} & human summarized & 300k     & F/NF       \\ \noalign{\hrule height 1.2pt}
		\end{tabular}
		\caption{Different MRC datasets. `OQA', `MC' refers to \textit{open question answering} and \textit{multiple-choice} respectively. \textit{datasize} is the whole data size regardless the train/dev/test. F and NF denotes whether the question is factoid or non-factoid.}\label{dataset}
	\end{table*}
	
	\section{Related Work}
	
	The MRC system from the NLP community could date back to 1990s when \citet{hirschman1999deep} proposed a bag-of-words method that could give the answer to arbitrary text input. However, MCTest \citep{richardson2013mctest} is widely recognized as the first dataset that we could build machine learning systems on it. Since the proposal of MCTest, there are more and more MRC datasets curated to facilitate MRC development. Table \ref{dataset} shows an overview of these datasets and we divided them into three categories based on the answer type:

	\textbf{Multiple-Choices} is the standard type of reading comprehension that contains several candidates. MCTest is a canonical multiple-choice dataset where each question is combined with 4 options. The MCTest is curated by experts and restricted to the concepts that a 7-year-old is expected to understand. Bioprocess \citep{berant2014modeling} is another multiple-choice MRC dataset where the paragraph describing a biological process, and the goal is to answer questions that require an understanding of the relations between entities and events in the process. Other multiple choices MRC datasets including MCScript \citep{Ostermann2018MCScriptAN} that requires the system to understand the script of daily events, and RACE \citep{Lai2017RACELR} where the questions are collected from the English exams of Chinese students.

	\textbf{Cloze} is another type of MRC test in which some key points in the text are removed and should be filled given the contexts \citep{taylor1953cloze}. Cloze could be deemed as complementary to multiple-choice reading comprehension for its reduced redundancy in text \citep{spolsky1969reduced}.
	\citet{Hermann2015TeachingMT} use the article of CNN/Daily Mail as context, and blank out the entities in the summaries as the questions. Children's Book Test (CBT) \citep{Hill2016TheGP} is another automatic generated cloze data. In CBT, a random entity was removed from a sentence and should be predicted given the previous 20 sentences. Clicr \citep{vsuster2018clicr} is a medical domain cloze style data containing clinical case reports with about 100k gap-filling queries.
	
	\textbf{Open question answering} is the dominant data type of current MRC where there are no options and the system must generate the answer. Most models in this types of dataset sometimes resort to the extractive strategy. SQuAD \citep{rajpurkar2016squad} is built upon Wikipedia where the context is a Wikipedia paragraph and the questions and answers were crowdsourced. SQuAD2.0 \citep{Rajpurkar2018KnowWY} is an extension to SQuAD that each document was given some questions that could not be answered. NewsQA \citep{Trischler2017NewsQAAM} is based on CNN/Daily Mail, the answers and questions are generated by different people to solicit exploratory questions that require reasoning. 
	%    A similar strategy was adopted in DuoRC \citep{Saha2018DuoRCTC} where the question and answer are given based on two versions of the same movie description.
	NaturalQuestions \citep{kwiatkowski2019natural} is a Wikipedia based dataset focusing on factoid questions. In SearchQA \citep{dunn2017searchqa} and MARCO \citep{Nguyen2016MSMA} the documents were collected from search engine. 
	
	ReCO is also related to another Chinese MRC dataset such as DRCD \citep{Shao2018DRCDAC}, CMRC \citep{cmrc2017dataset} and DuReader \citep{He2018DuReaderAC}. Specifically, DuReader also contains yes/no questions. However, it only contains a small portion (8\%) of the yes/no questions, and only the whole documents are provided as context, which contains much more irrelevant information, or may directly answer the questions without deep reasoning of the evidence.

	Compared with other datasets, ReCO is an opinion based MRC dataset focusing on the yes/no/uncertain questions. The questions and context are obtained from real-world queries and web pages which shows diversity in domains. Besides, the context passage in ReCO is very short evidence and in most cases, deep inference skills such as analogue, logical reasoning are required to answer the question. ReCO is also related to recognizing textual entailment (\textbf{RTE}) where the task is to determine whether there exits entailment/neutral/contradiction relation between two sentences, such as SNLI \citep{snli:emnlp2015} and MNLI \citep{williams2017broad}. However, entailment is a more narrow concept that the truth of hypothesis strictly requires the truth of the premise, whereas in ReCO the \textit{premise} is the evidence and \textit{hypothesis} is the question, so the inference between them contains much broader concepts such as deduction, induction, and abduction.

	\section{Data Collection}
	
\begin{figure}[t!]
		\centering
		\includegraphics[width=0.8\linewidth]{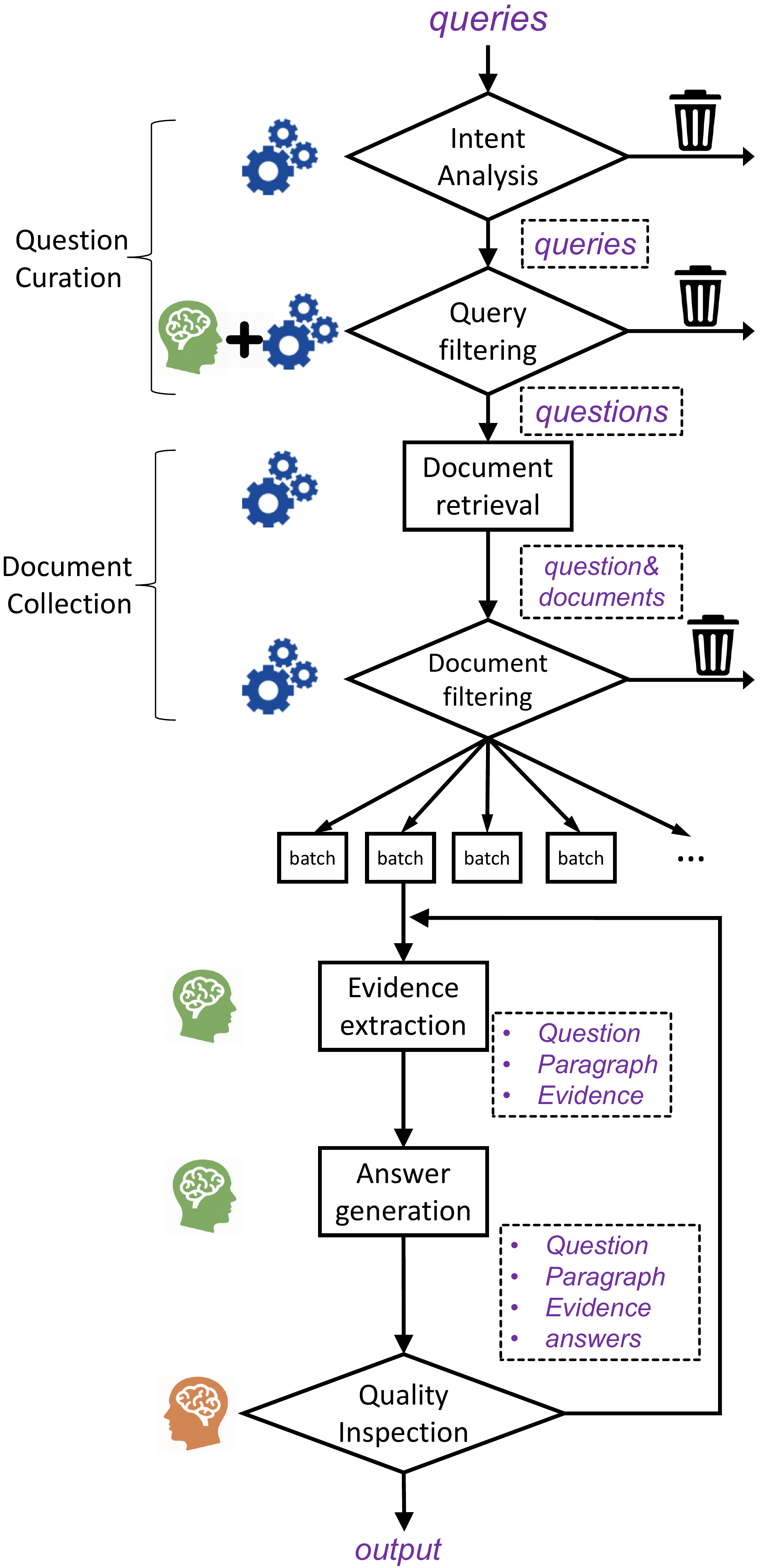}
		\caption{The data collection process of ReCO. \annotator  refers the annotator, \checker refers the authoritative checker, \machine refers to off-the-shelf system or machine learning models.}\label{fig_process}
	\end{figure}

	The data collection process includes a query selection, passage retrieval, passage filtering, evidence extraction, and answer generation process. Rigorous inspections are applied to ensure the quality and difficulty of the datasets. The conceptual scheme is illustrated in Figure \ref{fig_process}.
	
	\subsection{Qustion Curation}
	\textbf{Intent Analysis}: First of all, we sample 10 million queries issued to Sogou Search Engines. Next, we use the off-the-shelf intent analysis system to determine whether the query is a valid question. Then we drop some queries that contain sex, violence and other inappropriate content. These two processes exclude nearly 95 percent queries. 
	
	\noindent
	\textbf{Query Filtering}: Given the filtered queries, we build a simple symbolic feature based machine learning system to determine whether the question could be answered by yes or no. The features we use are whether it contains `Could I', `whether', `Is there any', etc. This simple system is very effective that only a small fraction of questions do not present the yes/no query intent.
	
	After the intent analysis and query filtering process, we obtain the original questions, although some invalid questions may pass through the above filtering processes. The next several steps could further reject some of these questions to make invalid questions as less as possible. 
	
	\subsection{Document Collection}
	\textbf{Document Retrieval}: we use the off-the-shelf Sogou search engine to retrieve 10 pages for each question, and then extract the body content of each page. The main focus of ReCO is text understanding but not fact seeking, so we did not filter out the pages from the forums or community sites where the answers may be subjective.
	
	\noindent
	\textbf{Document Filtering}: is proposed to prevent the retrieved documents containing some trivial answers that perfectly match the question. For example, if the question is `Can pregnant women eat celery?', it would be meaningless to give the candidate document which contains `pregnant women can eat celery'. We use some word-based rules to remove the documents that contain significant surface overlap with the question.
	
	\subsection{Evidence \& Answer Curation}
	We first randomly divided the question-documents pairs into a lot of batches with each batch contains 5k samples. Then each batch is annotated by a single annotator by the following processes.
	
	\noindent
	\textbf{Evidence Extraction:} given a question and its relevant documents, we ask the annotator to extract the snippet from the document as the evidence. And the document containing the evidence is the context paragraph. There are four principles for this extraction: 
	\begin{itemize}
		\item  The evidence should be self-contained to answer the question and as short as possible.
		\item  If multiple support evidence could answer the question, select the most implicit one requiring deeper inference.
		\item If there is no evidence in the document that could answer the question, select the most relevant passage.
		\item If a question could not be answered by yes/no/undefined, it should be rejected.
	\end{itemize}
	
	The first principle is introduced so that the extracted evidence should contain less irrelevant information to the question, and therefore bypass the answer selection errors which is the bottleneck in some other datasets. The second principle ensures the difficulty of the extracted passage, enabling deeper reasoning of the text.
	
\begin{figure}
		\centering
		\subfigure[negative answers]{
			\begin{minipage}[b]{0.465\linewidth}
				\includegraphics[width=0.9\linewidth]{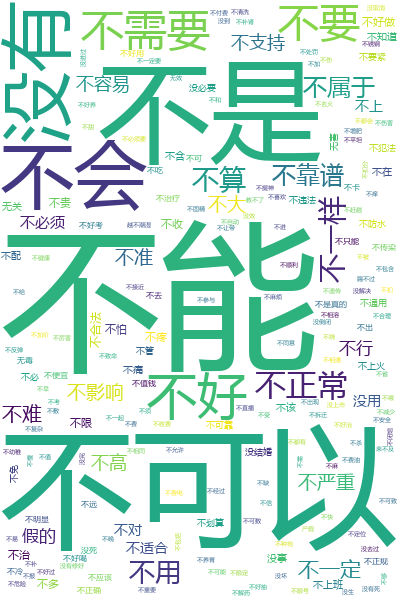}
			\end{minipage}
		}
		\subfigure[positive answers]{
			\begin{minipage}[b]{0.465\linewidth}
				\includegraphics[width=0.9\linewidth]{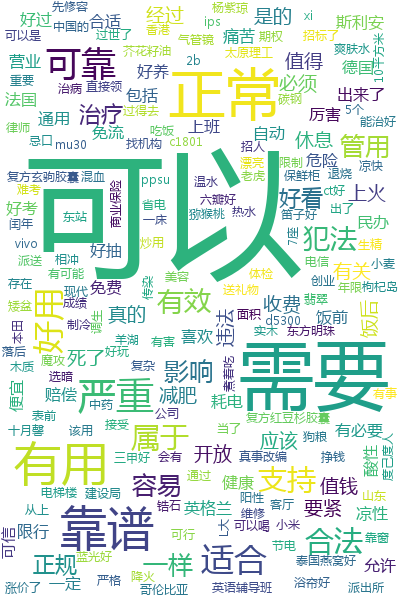}
			\end{minipage}
		}        
		\caption{Wordcloud of candidate answers. The common positive answers are \textit{can, need, yes, useful}. And the most common negative answers are \textit{can't, unable, not, wouldn't,}.\label{fig_pos_neg}}
	\end{figure}

	\noindent
	\textbf{Answer Generation:} 
	The annotator was asked to give abstractive candidates to the answer after the evidence extraction process. It should contain a positive one such as `Can' and a negative one such as `Cannot', and an undefined one if the question could not be answered by the passage\footnote{as we may remove the valid documents in the document filtering process, or the question is too unusual to get a good answer. In this case, the passage is selected as the most relevant snippet in the documents to the question.}. The answer candidates are summarized by the annotator that it may not be the words in the original evidence or question which are shown in Figure \ref{fig_pos_neg}.
	%    Figure \ref{fig_pos_neg} shows some frequent words of the answers.
	
	%    
	
	\subsection{Quality Inspection}
	After the above processes, we obtain a lot of batches, each batch is examined by the expert checkers who are expert at our domain and fully understand the demanding quality of the dataset. There are four key rules for the expert checker to determine whether a sample is false:
	\begin{itemize}    
		\item[$\star$] The answer is incorrect.
		\item[$\star$] The question is blurred, or it could not be answered by yes/no/undefine.
		\item[$\circ$] The evidence has much irrelevant information.
		\item[$\circ$] The question is too easy given the evidence.
	\end{itemize}
	The rules with $\star$ are the strict rules that the instance is absolute false. The rules with $\circ$ are the loose rules that this instance is only \textit{half wrong} and only account 0.5 in the final error summation.
	
	For each batch, we randomly sample a fraction of instances and send it to the expert checker. The checker examines the quality of these data based on the above rules. If the accuracy of a batch is higher than 0.95, it is passed and accepted to the final set. Otherwise, this batch is rejected and pushed back to the evidence extraction process of the corresponding annotator and relabeled again.
	
	%    50 annotators and 4 expert checkers were involved in the labeling process. 0.5 RMB was paid to each annotator for each valid samples. The labeling process takes about 4 weeks and finally we obtain nearly 300k valid samples with total 233,333 RMB.
	
	\subsection{Test Data Collection}
	The test data requires higher quality compared with the training set for its evaluation usage. After the document filtering process, we sent each sample to two annotators and annotated independently. If the answer is the same across two annotators, it was sent to the third annotator to select the evidence provided by the preceding two annotators.
	
	Finally, we obtain 280,000 training data and 20,000 testing data. The average length of paragraph, evidence and question is 924.5 characters, 87.1 characters and 10.6 characters respectively. The ratio of positive, negative, undefined questions is about 5:4:1.
	
	\section{Dataset Analysis}
	To understand the properties of ReCO, we sample 200 instances to analyze three aspects of ReCO: (\romannumeral1) The diversity of the question domains. (\romannumeral2) The domains of the evidence. (\romannumeral3) The reasoning skills required to answer the questions.
	
	\subsection{Question Domains}
	The diversity of question domains could somewhat reflect the world knowledge coverage of the data. We divided the questions into 5 categories: (1) Health: about disease, foods, exercise, etc. (2) SciTech: including science, technique, tools, etc. (3) Society: questions about legal provisions, stipulations, education, etc. (4) Life: questions about life such as public transport information, vacation, shopping, etc. (5) Culture: about literature, art, history, etc. In Table \ref{question_domains}, we can see that the question domain is varied. This is an advantage over some previous works such as SearchQA, or NarrativeQA that the question is focused on specific domain. The diversity in ReCO question makes it a comprehensive dataset containing many aspects of the world knowledge.
	\begin{table}[]
		\centering
		\includegraphics[width=0.9\linewidth]{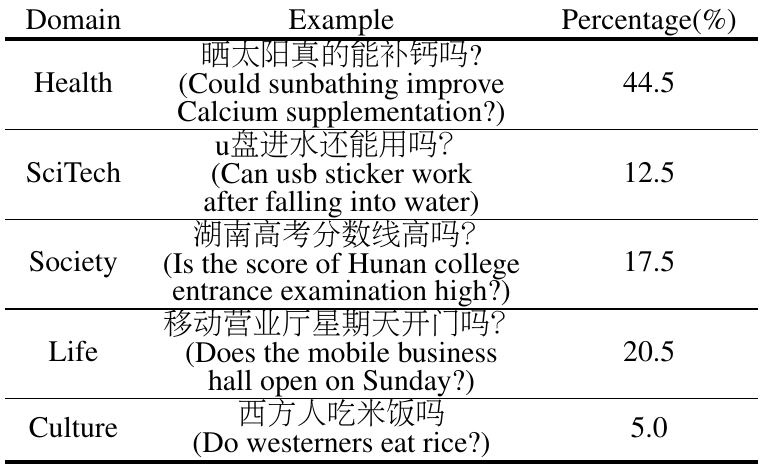}
		\caption{Question domains of ReCO.}\label{question_domains}
	\end{table}

	\subsection{Evidence Domains}
	The documents in ReCO are derived from the search engine, so pages from every possible domain could be the candidate answer. We further analyze the diversity of the evidence domains. We coarsely divide the evidence into three categories based on its source document: (1) Vertical: sites providing authoritative information for specific domain, such as medical center, government legislation department, etc. (2) Forum: the online discussion site where people can hold conversations in the form of posted messages. (3) Others: including the evidence from other sources such as encyclopedia, news articles, etc. 
	
	In Table \ref{passage_domains}, we can see the evidence domain of ReCO is diverse, including both formal texts such as vertical sites articles and informal text such as forum discussions. This is an advantage over some other datasets of which the domains are limited to certain types, such as Wikipedia \citep{rajpurkar2016squad,kwiatkowski2019natural}, news articles \citep{Hermann2015TeachingMT} or children stories \citep{Hill2016TheGP}, which elude understanding the stylistic feature of text and hard to test the generalization ability of existing models. The diversity in passage domains is the prerequisite of literary form understanding which is a key point in reading comprehension \citep{snow2002reading,makhoul2015reading}.
	\begin{table}[]
		\centering
		\fontsize{8.6pt}{9pt}\selectfont
		\begin{tabular}{ccc}
			Domain     & Example                & Percentage(\%) \\ \noalign{\hrule height 1.15pt}        
			Vertical   & FDA, MamaBang          &     40.0       \\ \hline
			Forum      & Quora, Sogou Wenwen    &     44.5       \\ \hline
			Others       & Sogou Baike, People's daily         &      15.5      \\ \noalign{\hrule height 1.15pt}
		\end{tabular}
		\caption{Evidence Domains of ReCO. The classification is based on the source document of the evidence.}\label{passage_domains}
	\end{table}
	\begin{table*}[!h]
		\centering
		\includegraphics[width=1.0\linewidth]{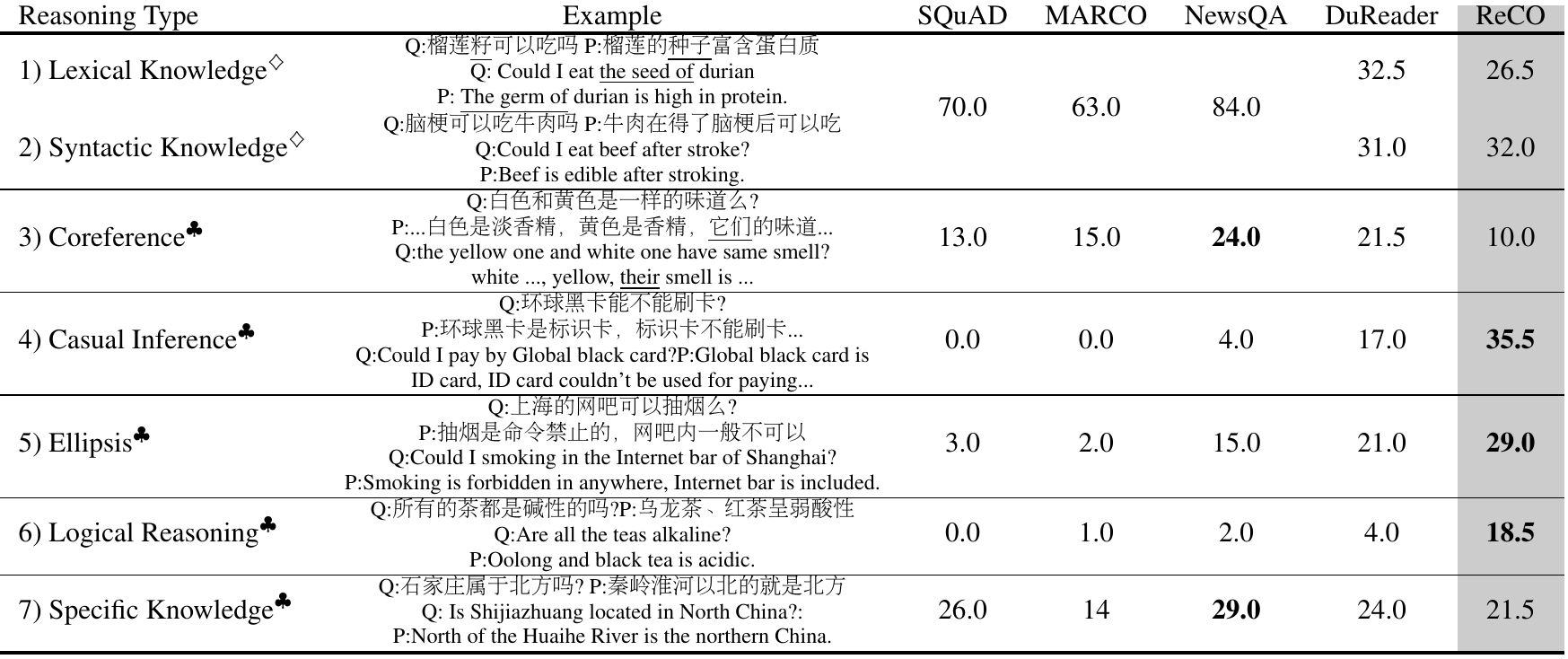}
		\caption{Frequencies (\%) of the shallow ($ ^\diamondsuit $) or deep($ ^\clubsuit $) reasoning skills required for MRC datasets. The results of SQuAD, MARCO and NewsQA are borrowed from \citet{sugawara2017evaluation}. We exclude the `undefined' questions because the evidence may not relate to the question. Note that some question requires more than one skill.}\label{reason_skill}
	\end{table*}

	\subsection{Reasoning Skills}
	To achieve natural language understanding, reasoning skills are the main desiderata of MRC datasets compared with other factors. To get a better understanding of reasoning skills required to answer the questions in ReCO, following previous works on reading comprehension \citep{mcnamara2009toward} and MRC \citep{sugawara2017evaluation,sugawara2018makes}, we identified 7 reasoning skills and classify them into shallow ($ ^\diamondsuit $) or deep($ ^\clubsuit $)\footnote{Note that this categorization is empirical and provisional without very solid theoretical background.}: 
	
	\textbf{1) Lexical Knowledge$ ^\diamondsuit $:} lexical information about the word, such as synonymy, hypernymy or morphology.
	
	\textbf{2) Syntactic Knowledge$ ^\diamondsuit $:} is the knowledge about sentence structure, such as part of speech, apposition or clause relation such as coordination or subordination, including relative clauses. 
	
	\textbf{3) Coreference$ ^\clubsuit $:} is a skill to track some objects, including anaphora and cataphora.
	
	\textbf{4) Casual Inference$ ^\clubsuit $:} is the knowledge about causality between the cause and effect, which are sometimes represented by events.
	
	%    \textbf{5) Analogue:} the indirect reference to something, such as metonymy and synecdoche.
	
	\textbf{5) Ellipsis$ ^\clubsuit $:} recognizing omitted information (argument, predicate, quantifier, place, etc).
	
	\textbf{6) Logical Reasoning$ ^\clubsuit $:} understanding the predicate logic such as negation, conditionals, quantifiers, transitivity.
	
	\textbf{7) Specific Knowledge$ ^\clubsuit $:} is the skills in specific domain, temporal, mathematical, spatiotemporal among others.
	
	The statistics of different reasoning skills are shown in Table \ref{reason_skill}. We can see that compared with other prevalent MRC datasets, a large amount of questions in ReCO requires deeper reasoning skills. Specifically, we found that many questions require casual inference and ellipsis which are very difficult for current systems. Reviewing the data collection process, we conclude two factors making the ReCO more difficult: 
	
	\textbf{Intrinsic}: In ReCO, we confine the answer type to be yes/or/uncertain, thus the question answering can be cast as an entailment recognition process. However, as the questions are real-world queries generated by users and the evidence is the extraction of the document retrieved by the search engine, there is no direct correspondence between the `\textit{premise}', i.e. the evidence, and the `\textit{hypothesis}'-the question. This is in contrast with many previous MRC datasets, where the questions are generated based the context documents, so there is a strong correlation between the question and context in previous datasets that sometimes trivial to answer. The information decoupling between the evidence and question in ReCO necessitate the deeper understanding of the textual inference.
	
	\textbf{Extrinsic}: the data collection process involves a rigorous quality inspection step that some data would be eliminated if it was too easy. Although this step may result in a large decrease of the data size, we believe it is indispensable to remove too easy samples given that the MRC systems are prone to be attacked by adversarial examples \citep{Jia2017AdversarialEF}, for which the simple pattern in the data is the main reason \citep{sugawara2018makes}.
	
	%    The goal of ReCO is to provide a testbed for natural language understanding system. 
	
	\section{Experiments}
	\subsection{Baselines}
	To evaluate the baseline performance, we consider three competitive models that perform very well in MRC and many other NLU applications:
	
	\begin{itemize} 
		\item  \textbf{BiDAF} \citep{Seo2016BidirectionalAF} is the very first deep learning model achieving remarkable performance on MRC. It is built upon LSTM and bi-directional attention was introduced to interact the question and the answer.  
		\item  \textbf{BiDAF$ ^* $}: In addition to BiDAF, ELMO \citep{Peters2018DeepCW} is introduced for word representation initialization. ELMO is an LSTM based bi-directional language model trained on unsupervised data which shows advantage compared to the word embedding methods.
		\item \textbf{BERT} \citep{devlin2018bert} is a recently proposed model that substantially advanced the state-of-the-art in many NLP tasks. It is based on Transformer \citep{Vaswani2017AttentionIA} and pre-trained in large unlabeled data, the objective contains a mask language modeling task and a next sentence prediction task.
	\end{itemize}
	
	The original BERT model is not designed for the multiple-choice problem, so we modified the architecture to take the candidate answers into account. Concretely, we concatenate the candidate answers after the sequence with a special \texttt{[CLS]} token. Then we use the BERT to represent the whole sequence. Finally, the output representations of the three \texttt{[CLS]} tokens are used to predict the three candidate answers probability. The loss is the cross entropy.

	\subsection{Common Setup}
	We use the ReCO training set to build sentencepiece \citep{kudo2018sentencepiece} tokenizer and set the vocabulary size to 35,000. For BiDAF, we adopt the same experimental setting with the original implementation. For BERT and ELMO, we use the openly released code\footnote{\url{https://github.com/pytorch/hub}}. In all experiments we set the batch size to 48 and run on 8 Nvidia V100 GPUs.

	Table \ref{table_result} shows the result of these models on ReCO. Compared with their performance on other MRC datasets, It is clear that the current best models still struggle to achieve a good result in ReCO, even though it is a simple three-category classification problem and a random system could achieve 1/3. Needless to say the input is the short noise-free evidence. If we fed the long document instead of the evidence to the model, the results drop a lot, which means the answer selection process also plays a key role in MRC.
	\begin{table}[]
		\centering
		\includegraphics[width=1.0\linewidth]{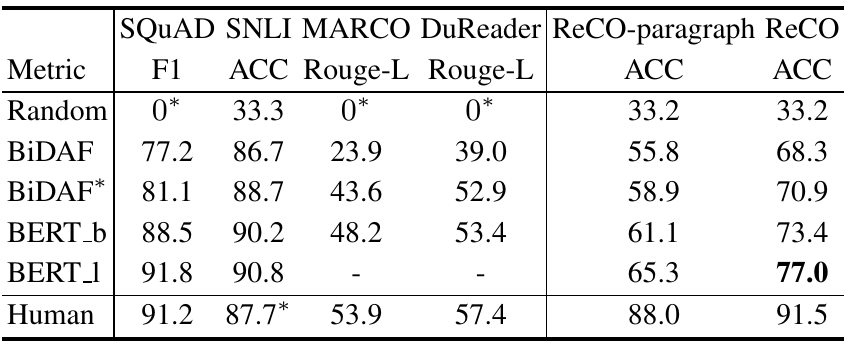}
		\caption{Result of different models on ReCO and other MRC datasets. $ ^* $ means the estimated results. ReCO-para denotes we use the original paragraph as the context. BERT$_{\_b|\_l} $ denotes BERT base or large model. The results of SQuAD, SNLI, DuReader and MARCO were derived from their leaderbord or paper.}\label{table_result}        
	\end{table}
	
\begin{figure}
		\centering
		\includegraphics[width=0.98\linewidth]{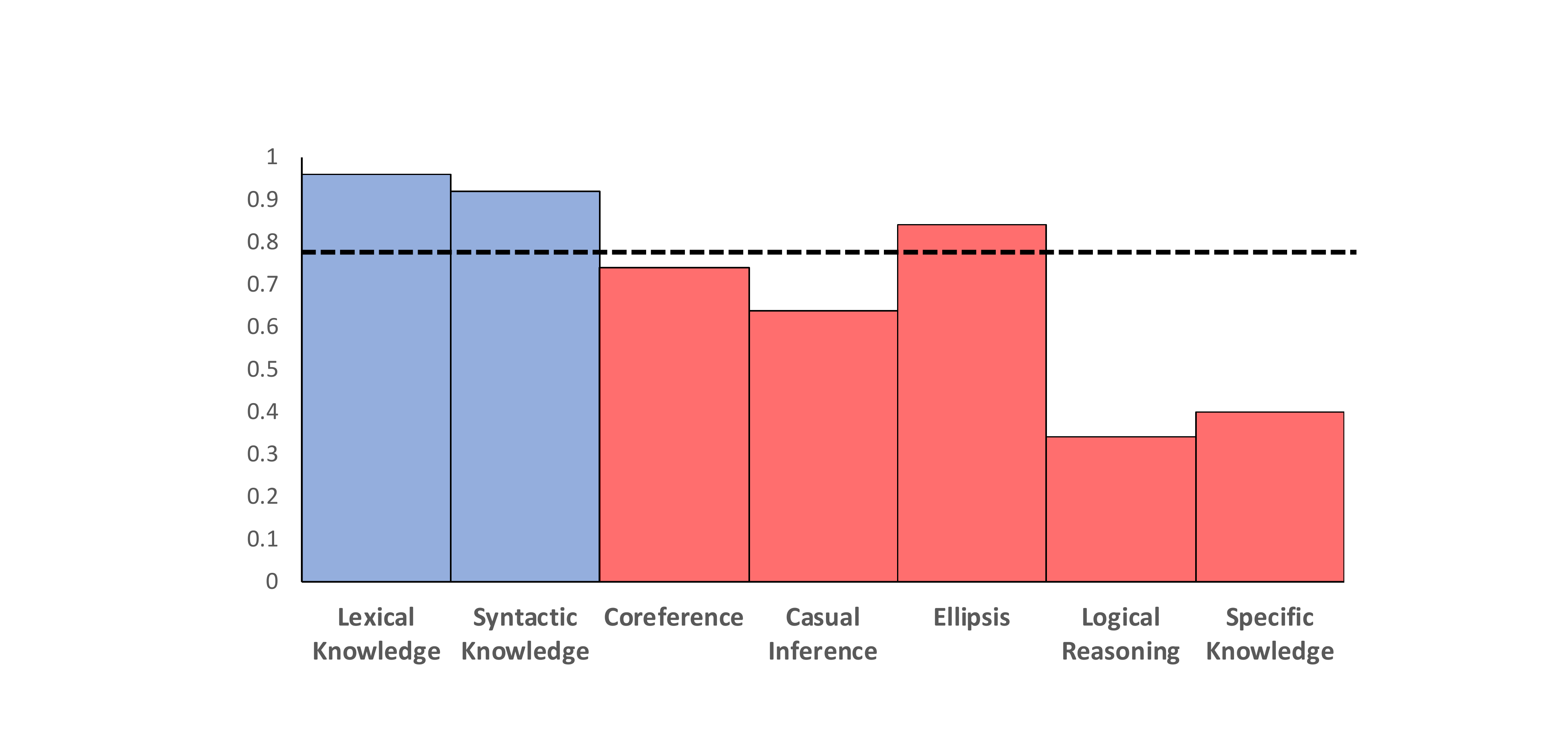}
		\caption{The accuracy of BERT large model on different types of questions based on reasoning skills in Table \ref{reason_skill}. The dotted line is the average accuracy.}\label{fig_error}
	\end{figure}

	We analysis the error of BERT large model on ReCO based on the inference skills in Table \ref{reason_skill}. The result is shown in Figure \ref{fig_error}. We can see that as Bert has been pre-trained with large unlabeled data, the lexical and syntactical knowledge has been well modeled and the corresponding accuracy is high. But the deeper inference skills, such as logical reasoning or specific knowledge, is not directly present in data so the performance of these questions is not satisfied. Incorporating more sophisticated knowledge, such as word sense \citep{sensebert} or knowledge base information may further benefit the model, which we leave for future study.
	
	\subsection{RI index}
	To understand the difficulty of a dataset, we proposed a relative improvement (\textbf{RI}) index:
	\begin{equation}\label{ri}
	\small
	\text{RI} = \frac{S_{\text{model}}-S_{\text{random}}}{S_{\text{human}}-S_{\text{random}}}
	\end{equation} 
	$ S_{\text{model}} $, $ S_{\text{random}} $ and $ S_{\text{human}} $ denote the score of best machine learning model, the score of a random system and the score of human beings, respectively. RI is measured by how much improvement the best machine learning models have achieved compared to how much improvement the human have achieved w.r.t. a random system. This index reflects the gap between the performance of current systems and human beings, and thus a criterion indicating the difficulty.
	
	\begin{figure}
		\centering
		\includegraphics[width=0.95\linewidth]{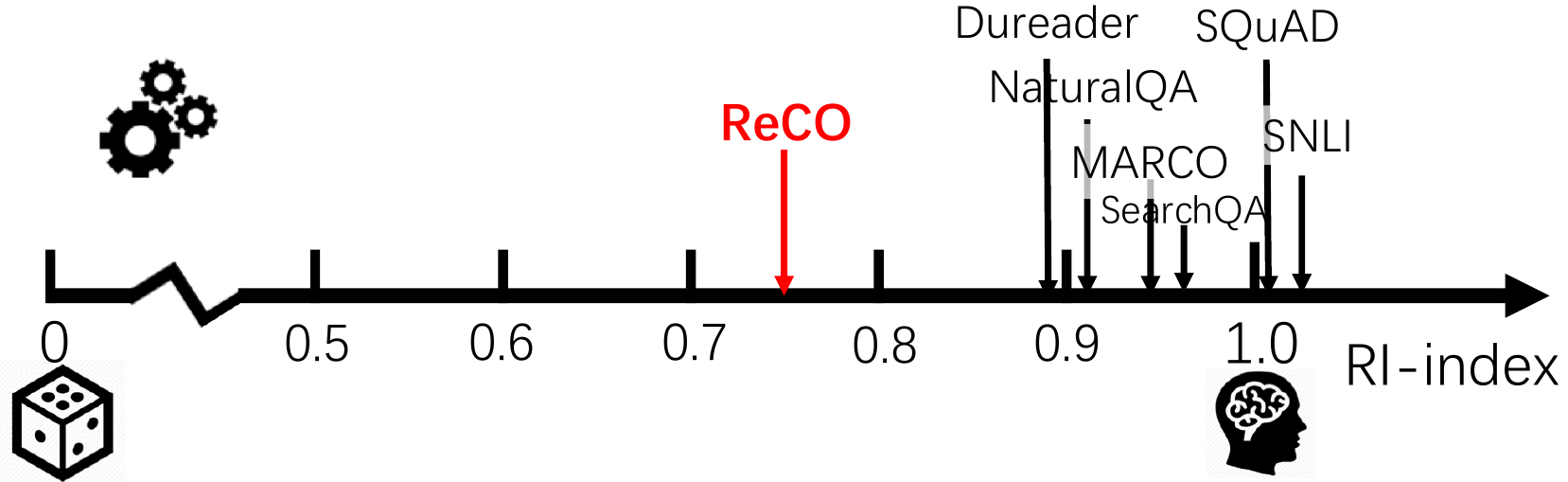}
		\caption{RI-index of different datasets. Results of other datasets obtained from their paper or leaderboard.}\label{fig_ri}
	\end{figure}

	Figure \ref{fig_ri} shows the RI indexes of different MRC datasets. It is clear that the machine learning models have achieved competitive results in most datasets, some of them even surpass human performance. But for ReCO, there is still a large headroom for machines to improve. 
	On the one hand, it reflects that the ReCO is a relatively hard task that the current model is still incompetent. On the other hand, most ReCO questions require deep reasoning skills, so new mechanisms should be introduced to the MRC models to achieve higher level inference, such as logical reasoning, etc.
	
	\section{Conclusion}
	This paper presents ReCO, a large scale opinion based Chinese reading comprehension dataset contains 300k questions. We use a very rigorous data inspection process to guarantee the quality of the data. ReCO contains short evidence which bypasses the answer selection error, and data analysis shows that most of the questions require deep reasoning skills. We develop a relative improvement index to measure the difficulty of the dataset. Experimental results and RI index demonstrate the difficulty of ReCO. Much more efforts should be made to filling the gap between machines and humans in this text understanding application.
	
	\section{Acknowledgments}
	We thank the anonymous reviewers for their insightful comments. We thank the Sinovation Ventures for hosting the AI Challenger 2018 competetions which introduce this dataset to the chinese NLP community.
	\small
	\bibliography{aaai2020}

\begin{thebibliography}{}

\bibitem[\protect\citeauthoryear{Alberti, Lee, and
  Collins}{2019}]{alberti2019bert}
Alberti, C.; Lee, K.; and Collins, M.
\newblock 2019.
\newblock A bert baseline for the natural questions.
\newblock {\em arXiv preprint arXiv:1901.08634}.

\bibitem[\protect\citeauthoryear{Berant \bgroup et al\mbox.\egroup
  }{2014}]{berant2014modeling}
Berant, J.; Srikumar, V.; Chen, P.-C.; Vander~Linden, A.; Harding, B.; Huang,
  B.; Clark, P.; and Manning, C.~D.
\newblock 2014.
\newblock Modeling biological processes for reading comprehension.
\newblock In {\em EMNLP}.

\bibitem[\protect\citeauthoryear{Bowman \bgroup et al\mbox.\egroup
  }{2015}]{snli:emnlp2015}
Bowman, S.~R.; Angeli, G.; Potts, C.; and Manning, C.~D.
\newblock 2015.
\newblock A large annotated corpus for learning natural language inference.
\newblock In {\em EMNLP 2015}.

\bibitem[\protect\citeauthoryear{Cui \bgroup et al\mbox.\egroup
  }{2018}]{cmrc2017dataset}
Cui, Y.; Liu, T.; Chen, Z.; Ma, W.; Wang, S.; and Hu, G.
\newblock 2018.
\newblock Dataset for the first evaluation on chinese machine reading
  comprehension.
\newblock In {\em LREC 2018}.

\bibitem[\protect\citeauthoryear{Devlin \bgroup et al\mbox.\egroup
  }{2018}]{devlin2018bert}
Devlin, J.; Chang, M.-W.; Lee, K.; and Toutanova, K.
\newblock 2018.
\newblock Bert: Pre-training of deep bidirectional transformers for language
  understanding.
\newblock {\em arXiv preprint arXiv:1810.04805}.

\bibitem[\protect\citeauthoryear{Dunn \bgroup et al\mbox.\egroup
  }{2017}]{dunn2017searchqa}
Dunn, M.; Sagun, L.; Higgins, M.; Guney, V.~U.; Cirik, V.; and Cho, K.
\newblock 2017.
\newblock Searchqa: A new q\&a dataset augmented with context from a search
  engine.
\newblock {\em arXiv preprint arXiv:1704.05179}.

\bibitem[\protect\citeauthoryear{He \bgroup et al\mbox.\egroup
  }{2018}]{He2018DuReaderAC}
He, W.; Liu, K.; Lyu, Y.; Zhao, S.; Xiao, X.; Liu, Y.; Wang, Y.; Wu, H.; She,
  Q.; Liu, X.; Wu, T.; and Wang, H.
\newblock 2018.
\newblock Dureader: a chinese machine reading comprehension dataset from
  real-world applications.
\newblock In {\em QA@ACL}.

\bibitem[\protect\citeauthoryear{Hermann \bgroup et al\mbox.\egroup
  }{2015}]{Hermann2015TeachingMT}
Hermann, K.~M.; Kocisk{\'y}, T.; Grefenstette, E.; Espeholt, L.; Kay, W.;
  Suleyman, M.; and Blunsom, P.
\newblock 2015.
\newblock Teaching machines to read and comprehend.
\newblock In {\em NIPS}.

\bibitem[\protect\citeauthoryear{Hill \bgroup et al\mbox.\egroup
  }{2016}]{Hill2016TheGP}
Hill, F.; Bordes, A.; Chopra, S.; and Weston, J.
\newblock 2016.
\newblock The goldilocks principle: Reading children's books with explicit
  memory representations.
\newblock {\em CoRR} abs/1511.02301.

\bibitem[\protect\citeauthoryear{Hirschman \bgroup et al\mbox.\egroup
  }{1999}]{hirschman1999deep}
Hirschman, L.; Light, M.; Breck, E.; and Burger, J.~D.
\newblock 1999.
\newblock Deep read: A reading comprehension system.
\newblock In {\em ACL},  325--332.
\newblock Association for Computational Linguistics.

\bibitem[\protect\citeauthoryear{Jia and Liang}{2017}]{Jia2017AdversarialEF}
Jia, R., and Liang, P.
\newblock 2017.
\newblock Adversarial examples for evaluating reading comprehension systems.
\newblock In {\em EMNLP}.

\bibitem[\protect\citeauthoryear{Kocisk{\'y} \bgroup et al\mbox.\egroup
  }{2018}]{Kocisk2018TheNR}
Kocisk{\'y}, T.; Schwarz, J.; Blunsom, P.; Dyer, C.; Hermann, K.~M.; Melis, G.;
  and Grefenstette, E.
\newblock 2018.
\newblock The narrativeqa reading comprehension challenge.
\newblock {\em TACL} 06:317--328.

\bibitem[\protect\citeauthoryear{Kudo and
  Richardson}{2018}]{kudo2018sentencepiece}
Kudo, T., and Richardson, J.
\newblock 2018.
\newblock Sentencepiece: A simple and language independent subword tokenizer
  and detokenizer for neural text processing.
\newblock {\em arXiv preprint arXiv:1808.06226}.

\bibitem[\protect\citeauthoryear{Kwiatkowski \bgroup et al\mbox.\egroup
  }{2019}]{kwiatkowski2019natural}
Kwiatkowski, T.; Palomaki, J.; Rhinehart, O.; Collins, M.; Parikh, A.; Alberti,
  C.; Epstein, D.; Polosukhin, I.; Kelcey, M.; Devlin, J.; et~al.
\newblock 2019.
\newblock Natural questions: a benchmark for question answering research.

\bibitem[\protect\citeauthoryear{Lai \bgroup et al\mbox.\egroup
  }{2017}]{Lai2017RACELR}
Lai, G.; Xie, Q.; Liu, H.; Yang, Y.; and Hovy, E.~H.
\newblock 2017.
\newblock Race: Large-scale reading comprehension dataset from examinations.
\newblock In {\em EMNLP}.

\bibitem[\protect\citeauthoryear{Levine \bgroup et al\mbox.\egroup
  }{2019}]{sensebert}
Levine, Y.; Lenz, B.; Dagan, O.; Padnos, D.; Sharir, O.; Shalev-Shwartz, S.;
  Shashua, A.; and Shoham, Y.
\newblock 2019.
\newblock Sensebert: Driving some sense into bert.
\newblock {\em CoRR}.

\bibitem[\protect\citeauthoryear{Levy \bgroup et al\mbox.\egroup
  }{2017}]{Levy2017ZeroShotRE}
Levy, O.; Seo, M.; Choi, E.; and Zettlemoyer, L.~S.
\newblock 2017.
\newblock Zero-shot relation extraction via reading comprehension.
\newblock In {\em CoNLL}.

\bibitem[\protect\citeauthoryear{Makhoul and
  Copti-Mshael}{2015}]{makhoul2015reading}
Makhoul, B., and Copti-Mshael, T.
\newblock 2015.
\newblock Reading comprehension as a function of text genre and presentation
  environment: comprehension of narrative and informational texts in a
  computer-assisted environment vs. print.
\newblock {\em Psychology} 6(08):1001.

\bibitem[\protect\citeauthoryear{McNamara and
  Magliano}{2009}]{mcnamara2009toward}
McNamara, D.~S., and Magliano, J.
\newblock 2009.
\newblock Toward a comprehensive model of comprehension.
\newblock {\em Psychology of learning and motivation} 51:297--384.

\bibitem[\protect\citeauthoryear{Min \bgroup et al\mbox.\egroup
  }{2018}]{min2018efficient}
Min, S.; Zhong, V.; Socher, R.; and Xiong, C.
\newblock 2018.
\newblock Efficient and robust question answering from minimal context over
  documents.
\newblock {\em arXiv preprint arXiv:1805.08092}.

\bibitem[\protect\citeauthoryear{Nguyen \bgroup et al\mbox.\egroup
  }{2016}]{Nguyen2016MSMA}
Nguyen, T.; Rosenberg, M.; Song, X.; Gao, J.; Tiwary, S.; Majumder, R.; and
  Deng, L.
\newblock 2016.
\newblock Ms marco: A human generated machine reading comprehension dataset.
\newblock {\em NIPS}.

\bibitem[\protect\citeauthoryear{Nishida \bgroup et al\mbox.\egroup
  }{2018}]{Nishida2018RetrieveandReadML}
Nishida, K.; Saito, I.; Otsuka, A.; Asano, H.; and Tomita, J.
\newblock 2018.
\newblock Retrieve-and-read: Multi-task learning of information retrieval and
  reading comprehension.
\newblock In {\em CIKM}.

\bibitem[\protect\citeauthoryear{Ostermann \bgroup et al\mbox.\egroup
  }{2018}]{Ostermann2018MCScriptAN}
Ostermann, S.; Modi, A.; Roth, M.~A.; Thater, S.; and Pinkal, M.
\newblock 2018.
\newblock Mcscript: A novel dataset for assessing machine comprehension using
  script knowledge.
\newblock {\em CoRR} abs/1803.05223.

\bibitem[\protect\citeauthoryear{Peters \bgroup et al\mbox.\egroup
  }{2018}]{Peters2018DeepCW}
Peters, M.~E.; Neumann, M.; Iyyer, M.; Gardner, M.; Clark, C.; Lee, K.; and
  Zettlemoyer, L.~S.
\newblock 2018.
\newblock Deep contextualized word representations.
\newblock In {\em NAACL-HLT}.

\bibitem[\protect\citeauthoryear{Rajpurkar \bgroup et al\mbox.\egroup
  }{2016}]{rajpurkar2016squad}
Rajpurkar, P.; Zhang, J.; Lopyrev, K.; and Liang, P.
\newblock 2016.
\newblock Squad: 100,000+ questions for machine comprehension of text.
\newblock In {\em EMNLP}.

\bibitem[\protect\citeauthoryear{Rajpurkar, Jia, and
  Liang}{2018}]{Rajpurkar2018KnowWY}
Rajpurkar, P.; Jia, R.; and Liang, P.
\newblock 2018.
\newblock Know what you don't know: Unanswerable questions for squad.
\newblock In {\em ACL}.

\bibitem[\protect\citeauthoryear{Ramamoorthy and
  Murugan}{2018}]{Ramamoorthy2018AnAS}
Ramamoorthy, S., and Murugan, S.
\newblock 2018.
\newblock An attentive sequence model for adverse drug event extraction from
  biomedical text.
\newblock {\em CoRR}.

\bibitem[\protect\citeauthoryear{Richardson, Burges, and
  Renshaw}{2013}]{richardson2013mctest}
Richardson, M.; Burges, C.~J.; and Renshaw, E.
\newblock 2013.
\newblock Mctest: A challenge dataset for the open-domain machine comprehension
  of text.
\newblock In {\em EMNLP}.

\bibitem[\protect\citeauthoryear{Seo \bgroup et al\mbox.\egroup
  }{2017}]{Seo2016BidirectionalAF}
Seo, M.~J.; Kembhavi, A.; Farhadi, A.; and Hajishirzi, H.
\newblock 2017.
\newblock Bidirectional attention flow for machine comprehension.
\newblock {\em ICLR}.

\bibitem[\protect\citeauthoryear{Shao \bgroup et al\mbox.\egroup
  }{2018}]{Shao2018DRCDAC}
Shao, C.-C.; Liu, T.; Lai, Y.; Tseng, Y.; and Tsai, S.
\newblock 2018.
\newblock Drcd: a chinese machine reading comprehension dataset.
\newblock {\em ArXiv}.

\bibitem[\protect\citeauthoryear{Snow}{2002}]{snow2002reading}
Snow, C.
\newblock 2002.
\newblock {\em Reading for understanding: Toward an R\&D program in reading
  comprehension}.
\newblock Rand Corporation.

\bibitem[\protect\citeauthoryear{Spolsky}{1969}]{spolsky1969reduced}
Spolsky, B.
\newblock 1969.
\newblock Reduced redundancy as a language testing tool.

\bibitem[\protect\citeauthoryear{Sugawara \bgroup et al\mbox.\egroup
  }{2017}]{sugawara2017evaluation}
Sugawara, S.; Kido, Y.; Yokono, H.; and Aizawa, A.
\newblock 2017.
\newblock Evaluation metrics for machine reading comprehension: Prerequisite
  skills and readability.
\newblock In {\em ACL}, volume~1,  806--817.

\bibitem[\protect\citeauthoryear{Sugawara \bgroup et al\mbox.\egroup
  }{2018}]{sugawara2018makes}
Sugawara, S.; Inui, K.; Sekine, S.; and Aizawa, A.
\newblock 2018.
\newblock What makes reading comprehension questions easier?
\newblock {\em EMNLP}.

\bibitem[\protect\citeauthoryear{{\v{S}}uster and
  Daelemans}{2018}]{vsuster2018clicr}
{\v{S}}uster, S., and Daelemans, W.
\newblock 2018.
\newblock Clicr: a dataset of clinical case reports for machine reading
  comprehension.
\newblock {\em NAACL}.

\bibitem[\protect\citeauthoryear{Taylor}{1953}]{taylor1953cloze}
Taylor, W.~L.
\newblock 1953.
\newblock “cloze procedure”: A new tool for measuring readability.
\newblock {\em Journalism Bulletin} 30(4):415--433.

\bibitem[\protect\citeauthoryear{Trischler \bgroup et al\mbox.\egroup
  }{2017}]{Trischler2017NewsQAAM}
Trischler, A.; Wang, T.; Yuan, X.; Harris, J.; Sordoni, A.; Bachman, P.; and
  Suleman, K.
\newblock 2017.
\newblock Newsqa: A machine comprehension dataset.
\newblock In {\em Rep4NLP@ACL}.

\bibitem[\protect\citeauthoryear{Vaswani \bgroup et al\mbox.\egroup
  }{2017}]{Vaswani2017AttentionIA}
Vaswani, A.; Shazeer, N.; Parmar, N.; Uszkoreit, J.; Jones, L.; Gomez, A.~N.;
  Kaiser, L.; and Polosukhin, I.
\newblock 2017.
\newblock Attention is all you need.
\newblock In {\em NIPS}.

\bibitem[\protect\citeauthoryear{Weissenborn, Wiese, and
  Seiffe}{2017}]{Weissenborn2017MakingNQ}
Weissenborn, D.; Wiese, G.; and Seiffe, L.
\newblock 2017.
\newblock Making neural qa as simple as possible but not simpler.
\newblock In {\em CoNLL}.

\bibitem[\protect\citeauthoryear{Williams, Nangia, and
  Bowman}{2017}]{williams2017broad}
Williams, A.; Nangia, N.; and Bowman, S.~R.
\newblock 2017.
\newblock A broad-coverage challenge corpus for sentence understanding through
  inference.
\newblock {\em arXiv preprint arXiv:1704.05426}.

\bibitem[\protect\citeauthoryear{Yu \bgroup et al\mbox.\egroup
  }{2018}]{Yu2018QANetCL}
Yu, A.~W.; Dohan, D.; Luong, M.-T.; Zhao, R.; Chen, K.; Norouzi, M.; and Le,
  Q.~V.
\newblock 2018.
\newblock Qanet: Combining local convolution with global self-attention for
  reading comprehension.
\newblock {\em ICLR}.

\end{thebibliography}
	\bibliographystyle{aaai}
	
\end{document}